\documentclass[accepted]{uai2025} 

\usepackage[american]{babel}

\usepackage{natbib} 
\usepackage{mathtools} 
\usepackage{booktabs} 
\usepackage{tikz} 

\usepackage{amsmath}
\usepackage{amssymb}
\usepackage{amsfonts}
\usepackage{amsthm}
\usepackage{nomencl}
\usepackage{subcaption}
\usepackage{svg}
\usepackage{soul}
\usepackage{booktabs}
\usepackage{xcolor}
\usepackage{hyperref}
\usepackage{url}
    
\newcommand{\startpara}[1]{{\vskip5pt\noindent{\bf #1.}}} 

\newcommand{\sectref}[1]{Section~\ref{#1}}
\newcommand{\figref}[1]{Figure~\ref{#1}}

\newcommand{\egref}[1]{Example~\ref{#1}}
\newcommand{\eqnref}[1]{Equation~\ref{#1}}
\newcommand{\thmref}[1]{Theorem~\ref{#1}}

\newcommand{\lemref}[1]{Lemma~\ref{#1}}

\newcommand{\apref}[1]{Appendix~\ref{#1}}

\newcommand{\egegref}[2]{Examples~\ref{#1} and \ref{#2}}

\newcommand{\figfigfigref}[3]{Figures~\ref{#1}, \ref{#2} and \ref{#3}}

\usepackage{ifthen}

\newtheorem{theorem}{Theorem}
\newtheorem{lemma}{Lemma}

\newcounter{exampcount}
\newenvironment{examp}
{\refstepcounter{exampcount}
\vskip6pt\noindent
{\it Example \arabic{exampcount}.}}
{\hfill$\blacksquare$\vskip6pt}

\def\Nset{\mathbb{N}}

\def\Rset{\mathbb{R}}
\def\Rsetgeq{\mathbb{R}_{\geq 0}}

\def\cM{{\mathcal{M}}}

\def\cA{{\mathcal{A}}}

\def\next{{\bigcirc}}
\def\until{{\mathsf{U}}}

\def\eventually{{\Diamond}}

\def\land{{\wedge}}
\def\lor{{\vee}}

\def\exp{{\mathbb{E}}}
\def\Sucq{{{\rm Suc}_q}}

\def\Rp{{R_{\mathsf{pg}}^{\otimes}}}
\def\Rh{{R_{\mathsf{hd}}^{\otimes}}}

\newcommand{\Rap}[1]{{R_{\mathsf{ap},#1}^{\otimes}}}
\newcommand{\Rah}[1]{{R_{\mathsf{ah},#1}^{\otimes}}}

\newcommand{\Vp}[1]{{V_{\mathsf{pg}}^{#1}}}
\newcommand{\Vh}[1]{{V_{\mathsf{hd}}^{#1}}}
\newcommand{\Vap}[2]{{V_{\mathsf{ap},#1}^{#2}}}
\newcommand{\Vah}[2]{{V_{\mathsf{ah},#1}^{#2}}}

\title{Adaptive Reward Design for Reinforcement Learning}
\author[1]{Minjae Kwon}
\author[1]{Ingy ElSayed-Aly}
\author[1]{Lu Feng}

\affil[1]{%
    The Department of Computer Science\\
    University of Virginia\\
    Charlottesville, VA 22904, USA
}

\begin{document}

\maketitle

\begin{abstract}
There is a surge of interest in using formal languages such as Linear Temporal Logic (LTL) to precisely and succinctly specify complex tasks and derive reward functions for Reinforcement Learning (RL). However, existing methods often assign sparse rewards (e.g., giving a reward of 1 only if a task is completed and 0 otherwise). By providing feedback solely upon task completion, these methods fail to encourage successful subtask completion. This is particularly problematic in environments with inherent uncertainty, where task completion may be unreliable despite progress on intermediate goals. To address this limitation, we propose a suite of reward functions that incentivize an RL agent to complete a task specified by an LTL formula as much as possible, and develop an adaptive reward shaping approach that dynamically updates reward functions during the learning process. Experimental results on a range of benchmark RL environments demonstrate that the proposed approach generally outperforms baselines, achieving earlier convergence to a better policy with higher expected return and task completion rate.
Code is available at 
\url{https://github.com/safe-autonomy-lab/AdaptiveRewardRL.git}.
\end{abstract}

%

\section{Introduction} \label{sec:intro} 

In reinforcement learning (RL), an agent's behavior is guided by reward functions, which are often difficult to specify manually when representing complex tasks. 
Alternatively, an RL agent can infer the intended reward from demonstrations~\cite{ng2000algorithms}, trajectory comparisons~\cite{wirth2017survey}, or human instructions~\cite{fu2018language}.
Recent years have seen a surge of interest in using formal languages such as Linear Temporal Logic (LTL) and finite automata to specify complex tasks and derive reward functions for RL (see the extensive list of related work in \sectref{sec:related}).
Nevertheless, existing methods often assign sparse rewards (e.g., giving a reward of 1 only if a task is completed and 0 otherwise).
Sparse rewards may necessitate hundreds of thousands of exploratory episodes for convergence to a quality policy.
Furthermore, many prior works are only compatible with specific RL algorithms tailored to their proposed reward structures, such as Q-learning for reward machines~\cite{camacho2019ltl}, modular DDPG~\cite{hasanbeig2020deep}, and 
hierarchical RL for reward machines~\cite{icarte2022reward}.

\emph{Reward shaping}~\cite{ng1999policy} is a paradigm where an agent receives some intermediate rewards as it gets closer to the goal and has shown to be helpful for RL algorithms to converge more quickly. 
Inspired by this idea, we develop a logic-based adaptive reward shaping approach in this work. 
We use the syntactically co-safe fragment of LTL to specify complex RL tasks, such as 
``the task is to touch red and green balls in strict order without touching other colors, then touch blue balls''. 
We then translate a co-safe LTL task specification into a deterministic finite automaton (DFA) and design reward functions that keeps track of the task completion status (i.e., a task is completed if an accepting state of the DFA has been reached). 

The principle underlying our approach is to assign intermediate rewards to an agent as it makes progress toward completing a task. A key challenge is how to measure the closeness to task completion. 
We adopt the notion of \emph{task progression} defined by~\cite{lacerda2019probabilistic}, which measures each DFA state's distance to accepting states. 
The smaller the distance, the higher degree of task progression. 
The distance is zero when the task is fully completed. 

Another challenge is what reward values to assign for various degrees of task progression. 
To this end, we design two different reward functions. 
The \emph{progression} reward function assigns rewards based on the reduced distance-to-acceptance values. 
The \emph{hybrid} reward function balances the progression reward and the penalty for self-loops (i.e., staying in the same DFA state). 
However, we find that optimal policies maximizing the expected return based on these reward functions may not necessarily lead to the best possible task progression.

To address this limitation, we develop an adaptive reward shaping approach that dynamically updates distance-to-acceptance values to reflect the actual difficulty of activating DFA transitions during the learning process. 
We then design two new reward functions, namely \emph{adaptive progression} and \emph{adaptive hybrid}, leveraging the updated distance-to-acceptance values.
We show that our approach can learn an optimal policy with the highest expected return and the best task progression within a finite number of updates. 

Finally, we evaluate the proposed approach on various discrete and continuous RL environments. 
Computational experiments show the compatibility of our approach with a wide range of RL algorithms. 
Results indicate our approach generally outperforms baselines, achieving earlier convergence to a better policy with higher expected return and task completion rate.

\subsection{Related Work} \label{sec:related} 

\cite{li2017reinforcement} presents one of the first works applying temporal logic to reward function design, assigning reward functions based on robustness degrees of satisfying truncated LTL formulas. \cite{de2019foundations} uses a fragment of LTL for finite traces (called LTL$_f$) to encode RL rewards. Several methods seek to learn optimal policies that maximize the probability of satisfying an LTL formula~\cite{hasanbeig2019reinforcement, bozkurt2020control, hasanbeig2020deep}. However, these methods assign sparse rewards for task completion and do not provide intermediate rewards for task progression.

There is a line of work on \emph{reward machines} (RMs), a type of finite state machine that takes labels representing environment abstractions as input and outputs reward functions. \cite{camacho2019ltl} shows that LTL and other regular languages can be automatically translated into RMs via the construction of DFAs. \cite{icarte2022reward} describes a collection of RL methods that exploit the RM structure, including \emph{Q-learning for reward machines} (QRM), \emph{counterfactual experiences for reward machines} (CRM), and \emph{hierarchical RL for reward machines} (HRM). These methods are augmented with potential-based reward shaping~\cite{ng1999policy}, where a potential function over RM states is computed to assign intermediate rewards. We adopt these methods (with reward shaping) as baselines for comparison in our experiments. As we will show in \sectref{sec:exp}, our approach generally outperforms baselines, providing more effective design of intermediate rewards for task progression.

\cite{jothimurugan2019composable} proposes a new specification language that can be translated into reward functions and later applies it for compositional RL in \cite{jothimurugan2021compositional}. These methods use a task monitor to track the degree of specification satisfaction and assign intermediate rewards. However, they require users to encode atomic predicates into quantitative values for reward assignment. In contrast, our approach automatically assigns intermediate rewards using DFA states' distance to acceptance values, eliminating the need for user-provided functions.

\cite{jiang2021temporal} presents a reward shaping framework for average-reward learning in continuing tasks. Their method automatically translates a LTL formula encoding domain knowledge into a function that provides additional reward throughout the learning process. 
This work has a different problem setup and thus is not directly comparable with our approach.

\cite{cai2023overcoming} proposes an approach that decomposes an LTL mission into sub-goal-reaching tasks solved in a distributed manner. The same authors also present a model-free RL method for minimally violating an infeasible LTL specification in \cite{cai2023learning}. Both works consider the assignment of intermediate rewards, but their definition of task progression requires additional information about the environment (e.g., geometric distance from each waypoint to the destination). In contrast, we define task progression based solely on the task specification, following~\cite{lacerda2019probabilistic}, which is a work on robotic planning with MDPs (but not RL).

\section{Background} \label{sec:background} 

\subsection{Reinforcement Learning} \label{sec:rl}

Consider an RL agent interacting with an environment modeled as an episodic \emph{Markov decision process} (MDP), where each learning episode terminates within a finite horizon $H$.
Formally, an MDP is denoted as a tuple $\cM=(S, s_0, A, T, R, \gamma, L)$ where $S$ is a set of states, $s_0 \in S$ is an initial state, $A$ is a set of actions, $T: S \times A \times S \to [0,1]$ is a probabilistic transition function, $R$ is a reward function, $\gamma \in [0,1]$ is a discount factor, and $L: S \to 2^{AP}$ is a labeling function with a set of atomic propositions $AP$. 
The reward function can be Markovian, denoted by $R: S \times A \times S \to \Rset$, 
or non-Markovian (i.e., history dependent), denoted by $R: (S \times A)^* \to \Rset$.
Both the transition function $T$ and the reward function $R$ are unknown to the agent.

At each timestep $t$, the agent selects an action $a_t$ given the current state $s_t$ and reward $r_t$. 
The environment transitions to a subsequent state $s_{t+1}$, determined by the probability distribution $T(\cdot | s_t, a_t)$, and yields a reward $r_{t+1}$.
A (memoryless) policy is defined as a mapping from states to probability distributions over actions,
denoted by $\pi: S \times A \to [0,1]$.
The agent seeks to learn an optimal policy that maximizes the expected return, 
represented by $\exp[\sum_{t=0}^{H-1} \gamma^t r_{t+1}]$.

\subsection{Co-Safe LTL Specifications} \label{sec:logic}

We utilize Linear Temporal Logic (LTL)~\cite{pnueli1981temporal}, which is a form of modal logic that augments propositional logic with temporal operators, to specify complex tasks for the robotic agent.
We focus on the syntactically co-safe LTL fragment, defined as follows. 
$$
\varphi := \alpha \ |\ \neg \alpha \ |\ \varphi_1 \land \varphi_2 \ |\ \varphi_1 \lor \varphi_2 \ |\ 
            \next \varphi \ |\ \varphi_1 \until \varphi_2 \ |\ \eventually \varphi
$$
where $\alpha \in AP$ is an atomic proposition, $\neg$ (negation), $\land$ (conjunction), and $\lor$ (disjunction) are Boolean operators, while $\next$ (next), $\until$ (until), and $\eventually$ (eventually) are temporal operators.  
Intuitively, $\next \varphi$ means that $\varphi$ has to hold in the next step; $\varphi_1 \until \varphi_2$ means that $\varphi_1$ has to hold at least until $\varphi_2$ becomes true; and $\eventually \varphi$ means that $\varphi$ becomes true at some time eventually. 
A co-safe LTL formula $\varphi$ can be converted into a DFA $\cA_\varphi$ accepting exactly the set of good prefixes for $\varphi$~\cite{kupferman2001model}. 
Formally, a DFA is denoted as a tuple $\cA_\varphi = (Q, q_0, Q_F, 2^{AP}, \delta)$, where
$Q$ is a finite set of states, $q_0$ is the initial state, $Q_F \subseteq Q$ is a set of accepting states, 
$2^{AP}$ is the alphabet, and $\delta: Q \times 2^{AP} \to Q$ is the transition function.

\begin{figure}[t]
     \centering
     \begin{subfigure}{0.4\columnwidth}
         \includegraphics[width=\textwidth]{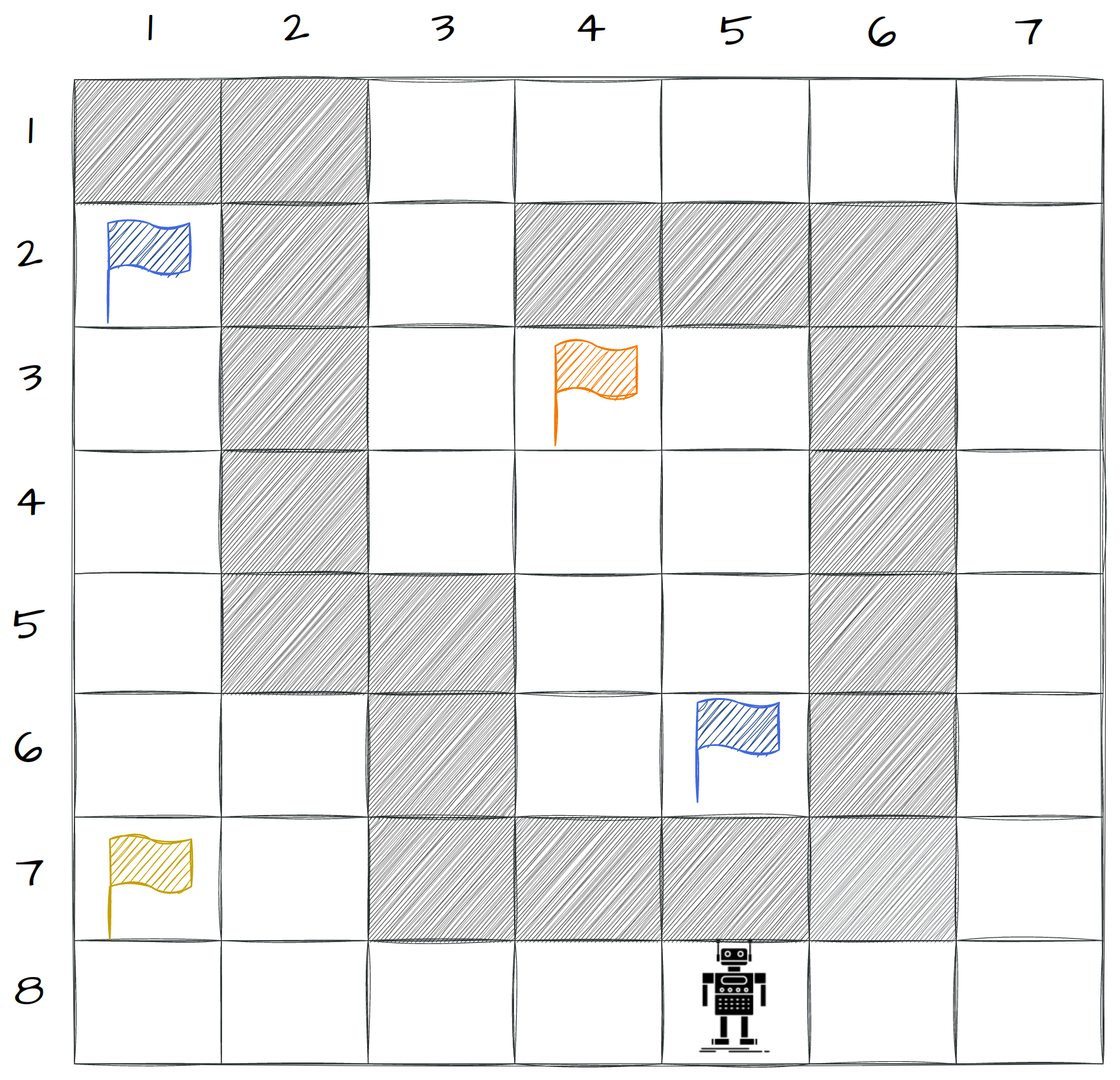}
         \caption{Gridworld}
         \label{fig:grid}
     \end{subfigure}
     \hfill
     \begin{subfigure}{0.55\columnwidth}
         \includegraphics[width=\textwidth]{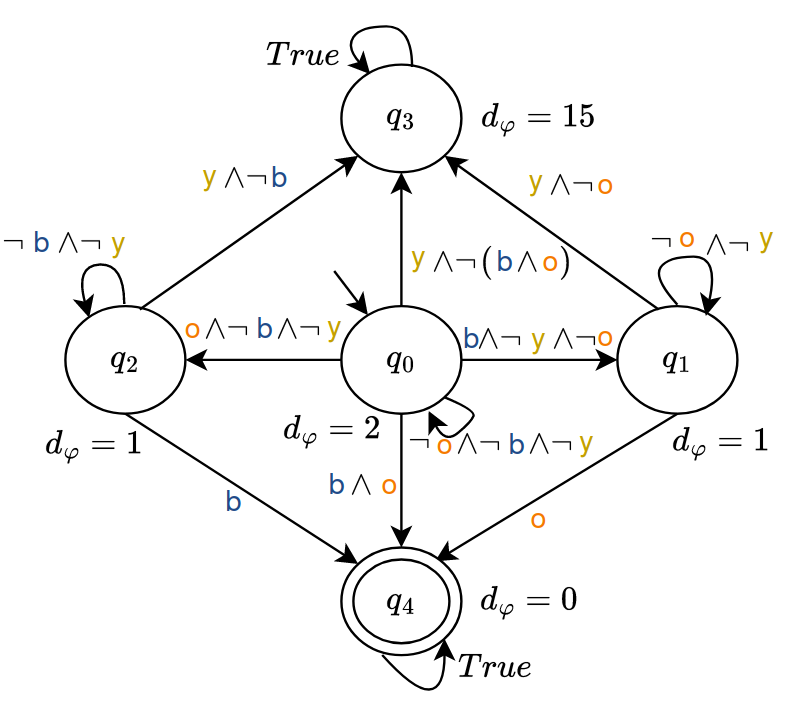}
         \caption{DFA $\cA_\varphi$}
         \label{fig:dfa}
     \end{subfigure}
\caption{Example gridworld and a DFA $\cA_\varphi$ for a co-safe LTL formula
$\varphi = (\neg y) \until ((o \land ((\neg y) \until b)) \lor (b \land ((\neg y) \until o)))$.}
\end{figure}

\begin{examp} \label{eg:dfa}
Consider a robot aiming to complete a task in a gridworld (\figref{fig:grid}).
The task is to collect an \emph{orange} flag and a \emph{blue} flag (in any order) while avoiding the \emph{yellow} flag. 
We describe this task using a co-safe LTL formula 
$\varphi = (\neg y) \until ((o \land ((\neg y) \until b)) \lor (b \land ((\neg y) \until o)))$, 
where $o$, $b$ and $y$ represent collecting \emph{orange}, \emph{blue} and \emph{yellow} flags, respectively. 
\figref{fig:dfa} shows the corresponding DFA $\cA_\varphi$, which has five states including the initial state $q_0$ depicted with an incoming arrow, a trap state $q_3$ from which no transitions to other states exist, and the accepting state $Q_F =\{q_4\}$ depicted with double circle.  
A transition is enabled when its labelled Boolean formula holds. 
Starting from the initial state $q_0$, a path ending in the accepting state $q_4$ represents a good prefix of satisfying $\varphi$, indicating that the task has been successfully completed. 
\end{examp}

\subsection{Task Progression}\label{sec:task}

We adopt the notion of ``task progression'' introduced in~\cite{lacerda2019probabilistic} to measure the degree to which a robotic task defined by a co-safe LTL formula $\varphi$ is completed. 

Given a DFA $\cA_\varphi = (Q, q_0, Q_F, 2^{AP}, \delta)$, let $\Sucq \subseteq Q$ be the set of successors of state $q$, and $|\delta_{q,q'}| \in \{0, \dots, 2^{|AP|}\}$ denote the number of possible transitions from $q$ to $q'$. We write $q \to^* q'$ if there is a path from $q$ to $q'$, and $q \not \to^* q'$ if $q'$ is not reachable from $q$.

The \emph{distance-to-acceptance function} $d_{\varphi}: Q \to \Rsetgeq$ is defined as:
\begin{equation}\label{eqn:distance}
    d_{\varphi}(q)= 
    \begin{cases}
        0 & \! \text {if } q \in Q_F \\ 
        \displaystyle \min_{q' \in \Sucq} d_{\varphi}(q') + h(q,q') 
            & \! \text {if } q \not \in Q_F \text {, } q \! \to^* Q_F \\ 
        |AP| \cdot |Q| & \! \text {otherwise }
    \end{cases} 
\end{equation}
where $h(q,q'):=\log_2 \left( \left\{ \frac{2^{|AP|}}{|\delta_{q,q'}|} \right\}\right)$ represents the difficulty of moving from $q$ to $q'$ in the DFA $\cA_\varphi$.

The \emph{progression function} $\rho_\varphi: Q \times Q \to \Rsetgeq$ between two states of $\cA_\varphi$ is defined as:
\begin{equation}\label{eqn:progression}
    \rho_\varphi (q, q') =   
    \begin{cases}
        \max \{0, d_\varphi(q) - d_\varphi(q') \} \!
            & \!\! \text{if} \, q' \in \Sucq \text{, } q' \! \not \to^* \! q \\
         0 \! & \!\! \text{otherwise }
    \end{cases}
\end{equation}
The first condition mandates $q' \not \to^* q$ to ensure that there is no cycle in the DFA with a non-zero progression value, which is crucial for the convergence of infinite sums of progression~\cite{lacerda2019probabilistic}.


\begin{examp}\label{eg:distance}
In the DFA $\cA_\varphi$ (\figref{fig:dfa}), 
the distance-to-acceptance values of the trap state $q_3$ and the accepting state $q_4$ 
is $d_\varphi(q_3)= 3 \times 5 = 15$ and $d_\varphi(q_4)=0$, respectively. 
Applying \eqnref{eqn:distance} recursively yields 
$d_\varphi(q_0)=2$, $d_\varphi(q_1)=1$, and $d_\varphi(q_2)=1$. 
The progression from the initial state $q_0$ to $q_1$ is 
$\rho_\varphi(q_0, q_1) = \max \{0, d_\varphi(q_0) - d_\varphi(q_1) \} = 1$, indicating that a positive task progression has been made.
\end{examp}

\section{Problem Formulation} \label{sec:problem} 

The objective of this work is to create reward functions that encourage an RL agent to achieve the best possible progression in accomplishing a task specified by a co-safe LTL formula $\varphi$.
To this end, we define a product MDP $\cM^\otimes$ that augments the environment MDP $\cM$ with information about the task specification $\varphi$. 

\startpara{Product MDP}
Given an episodic MDP $\cM=(S, s_0, A, T, R, \gamma, L)$ and a DFA $\cA_\varphi = (Q, q_0, Q_F, 2^{AP}, \delta)$,
the product MDP is defined as
$\cM^\otimes = \cM \otimes \cA_\varphi = (S^\otimes, s_0^\otimes, A, T^\otimes, R^\otimes, \gamma, AP, L^\otimes)$, where $S^\otimes = S \times Q$, $s_0^\otimes = \langle s_0, \delta(q_0, L(s_0))\rangle$,
$L^\otimes(\langle s,q \rangle) = L(s)$,
\begin{equation*}
    T^{\otimes}\left(\langle s, q\rangle, a,\langle s', q'\rangle\right) = 
    \begin{cases}
        T(s, a, s') & \text{if } q'=\delta(q, L(s'))\\
        0 & \text {otherwise. }
    \end{cases}    
\end{equation*} 
This work focuses on designing Markovian reward functions $R^\otimes: S^\otimes \times A \times S^\otimes \to \Rset$ for the product MDP $\cM^\otimes$, whose projection onto $\cM$ yields non-Markovian reward functions.

In practice, the product MDP is built on-the-fly during learning.
At each timestep $t$, given the current state $\langle s_t, q_t\rangle$, an RL agent selects an action $a_t$ and transits to a successor state $\langle s_{t+1}, q_{t+1}\rangle$, 
where $s_{t+1}$ is given by the environment, sampling from the distribution $T(\cdot | s_t, a_t)$, 
and $q_{t+1}$ is derived from the DFA's transition function $\delta(q_t, L(s_{t+1}))$. 
The agent receives a reward $r_{t+1}$ determined by the reward function 
$R^\otimes\left(\langle s_t, q_t\rangle, a,\langle s_{t+1}, q_{t+1}\rangle\right)$.

An RL agent aims to learn an optimal policy that maximizes the expected return in the product MDP $\cM^\otimes$.
A learned memoryless policy for $\cM^\otimes$ equates to a finite-memory policy in the environment MDP $\cM$, 
denoted by $\pi: S \times Q \times A \to [0,1]$,
with the DFA states $Q$ delineating various modes.

\startpara{Task progression for a policy}
We define a partition of the state space of DFA $\cA_\varphi = (Q, q_0, Q_F, 2^{AP}, \delta)$ based on an ordering of distance-to-acceptance values. 
Let $B_0 = Q_F$ and
$B_i = \{q \in Q \setminus \bigcup_{j=0}^{i-1} B_j \ |\ d_{\varphi}(q) \text{ is minimal} \}$
for $i > 0$. 
The task progression for a policy $\pi$ of the product MDP, denoted by $b(\pi)$, is the lowest index of reachable partitioned sets $B_i$ from the initial state.
A value of $b(\pi)=0$ signifies the task has been successfully completed.
The best possible task progression across all feasible policies $\Pi$ in the product MDP is defined as $b^* = \min\{b(\pi) \,|\, \pi \in \Pi \}$.

\begin{examp}\label{eg:policies}
The state space of the DFA $\cA_\varphi$ shown in \figref{fig:dfa} can be partitioned into four sets: $B_0 = \{q_4\}$, $B_1 = \{q_1, q_2\}$, $B_2 = \{q_0\}$, and $B_3 = \{q_3\}$. 

Let $g_{i,j}$ denote a grid cell in row $i$ and column $j$ in the gridworld (\figref{fig:grid}).
The agent's initial location is $g_{8,5}$. 
Consider the following three candidate policies:
\begin{itemize}
    \item $\pi_1$: The agent takes 10 steps to collect the blue flag in $g_{2,1}$, avoiding the yellow flag, but fails to reach the orange flag within the 25-step episode timeout.
    \item $\pi_2$: The agent moves 16 steps to collect the orange flag and then moves 4 more steps to collect the blue flag in $g_{6,5}$. The task is completed. 
    \item $\pi_3$: The agent moves directly to the yellow flag in 5 steps. The task is failed and the episode ends. 
\end{itemize}
We have $b(\pi_1) = 1$ as DFA state $q_1 \in B_1$ is reached with policy $\pi_1$,
$b(\pi_2) = 0$ upon task completion, and 
$b(\pi_3) = 2$ due to a direct transition from initial state $q_0 \in B_2$ to trap state $q_3 \in B_3$. 
The best possible task progression across all policies is $b^* = b(\pi_2) = 0$. 
\end{examp}

\startpara{Problem} 
This work aims to solve the following problem:
Given an episodic MDP $\cM$ with unknown transition and reward functions, along with a DFA $\cA_\varphi$ representing a co-safe LTL task specification $\varphi$, the objective is to construct a Markovian reward function $R^\otimes$ for the product MDP $\cM^\otimes = \cM \otimes \cA_\varphi$. This reward function should be designed such that an optimal policy $\pi^*$, learned by an RL agent via maximizing the expected return, also achieves the best possible task progression, that is, $b^* = b(\pi^*)$.

\section{Approach} \label{sec:approach} 

To solve this problem, we design two reward functions that incentivize an RL agent to improve the task progression (cf. \sectref{sec:reward}), and develop an adaptive reward shaping approach that dynamically updates these reward functions during the learning process (cf. \sectref{sec:adapt}).

\subsection{Basic Reward Functions} \label{sec:reward}

\startpara{Progression reward function}
First, we propose a \emph{progression reward function} based on the task progression function defined in \eqnref{eqn:progression}, representing the degree of reduction in distance-to-acceptance values.

\begin{align} \label{eqn:rp}
    & \Rp \left(\langle s, q\rangle, a,\langle s', q'\rangle\right) 
     = \rho_\varphi (q, q') \nonumber \\
    & =   
    \begin{cases}
        \max \{0, d_\varphi(q) - d_\varphi(q') \} 
            & \text{if} \, q' \in \Sucq \text{, } q' \! \not \to^* \! q \\
         0 & \text{otherwise }
    \end{cases}
\end{align}

\begin{examp} \label{eg:rp}
Assuming a deterministic environment for the gridworld shown in \figref{fig:grid}, 
the MDP has a discount factor of $\gamma = 0.9$.
We calculate the expected returns for policies from \egref{eg:policies} using the progression reward function. 
$\Vp{\pi_1}(s_0^\otimes) = 0.9^9 \approx 0.39$,
$\Vp{\pi_2}(s_0^\otimes) = 0.9^{15} + 0.9^{19} \approx 0.34$, and 
$\Vp{\pi_3}(s_0^\otimes) = 0$. 
Among these policies, $\pi_1$ yields the highest expected return, yet it fails to achieve the best possible task progression, as $b(\pi_1) = 1 > b^*=0$.  
\end{examp}

\startpara{Hybrid reward function}
The progression reward function rewards only transitions that progress toward acceptance, without penalizing those that stay in the same DFA state. 
To address this issue, we define a \emph{hybrid reward function}: 
\begin{equation} \label{eqn:rh}
    \Rh \left(\langle s, q\rangle, a,\langle s', q'\rangle\right) =   
    \begin{cases}
         \eta \cdot - d_\varphi(q) & \text {if } q = q' \\
         (1-\eta) \cdot \rho_\varphi (q, q') & \text {otherwise }
    \end{cases}
\end{equation}
where $\eta \in [0,1]$ balances the trade-offs between penalties and progression rewards. 

\begin{examp} \label{eg:rh}
We calculate the expected returns of policies in \egref{eg:policies} using the hybrid reward function (with $\eta = 0.1$). 
$\Vh{\pi_1}(s_0^\otimes) \approx -1.15$,
$\Vh{\pi_2}(s_0^\otimes) \approx -1.33$, and 
$\Vh{\pi_3}(s_0^\otimes) \approx -0.69$.
Although $\pi_3$ yields the highest expected return, 
it falls short in the task progression with $b(\pi_3) = 2$.
Increasing $\eta$ emphasizes penalties without altering the optimal policy in this example. Conversely, reducing $\eta$ moves closer to the progression reward function, especially when $\eta=0$.
\end{examp}

\subsection{Adaptive Reward Shaping} \label{sec:adapt}

While reward functions defined in \sectref{sec:reward} motivate an RL agent to complete a task specified by a co-safe LTL formula, \egegref{eg:rp}{eg:rh} show that the learned optimal policies that maximize the expected return do not achieve the best possible task progression.
A potential reason is that the distance-to-acceptance function $d_\varphi$, as defined in \eqnref{eqn:distance}, may not precisely reflect the difficulty of activating desired DFA transitions within a specific environment.
To tackle this limitation, we develop an adaptive reward shaping approach that dynamically updates distance-to-acceptance values and reward functions during the learning process.

\startpara{Updating distance-to-acceptance values}
After every $N$ learning episodes, with $N$ being a hyperparameter, we evaluate the average success rate of task completion. An episode is deemed successful if it concludes in an accepting state of the DFA $\cA_\varphi$.
If the average success rate falls below a predefined threshold $\lambda$, we proceed to update the distance-to-acceptance values accordingly.

We derive initial values $d_\varphi^0(q)$ for each DFA state $q \in Q$ from \eqnref{eqn:distance}.
The distance-to-acceptance values for the $k$-th update round are calculated recursively as follows:
\begin{equation} \label{eqn:updateD}
        d_\varphi^{k}(q) =   
    \begin{cases}
         d_\varphi^{k-1}(q) + \theta & \text{if } q \in B_i, \forall i \ge b_k \\
         d_\varphi^{k-1}(q) & \text {otherwise }
    \end{cases}
\end{equation}
where $b_k$ is the task progression of the optimal policy learned after $k \cdot N$ episodes, and $\theta$ is a hyperparameter, also used later in \eqnref{eqn:rah}, requiring that $\theta > 1$.

\begin{examp} \label{eg:updateD}
We have $d^0_{\varphi}(q_0) = 2$, $d^0_{\varphi}(q_1) = d^0_{\varphi}(q_2) = 1$, 
$d^0_{\varphi}(q_3) = 15$, and $d^0_{\varphi}(q_4) = 0$ following \egref{eg:distance}. 
Suppose $\pi_1$ is the optimal policy learned after the first $N$ episodes and thus $b_1 = 1$. 
Let $\theta = 100$. For states in $B_1 \cup B_2 \cup B_3 = \{q_0, q_1, q_2, q_3\}$, We update their distance-to-acceptance values as follows: $d^1_{\varphi}(q_1) = d^1_{\varphi}(q_2) = 101$,
$d^1_{\varphi}(q_0) = 102$, and $d^1_{\varphi}(q_3) = 115$.
For state $q_4 \in B_0$, we retain its distance-to-acceptance value as $d^1_{\varphi}(q_4) = 0$.  
\end{examp} 


Note that \eqnref{eqn:updateD} does not alter the order of distance-to-acceptance values, so the DFA state partitions $\{B_i\}$ remain unchanged. 
We present two new reward functions that leverage the updated distance-to-acceptance values as follows.

\startpara{Adaptive progression reward function}
Given the updated distance-to-acceptance values $d^k_{\varphi}(q)$, we apply the progression function defined in \eqnref{eqn:progression} and obtain
\begin{equation}\label{eqn:updateP}
    \rho^k_\varphi (q, q') =   
    \begin{cases}
        \max \{0, d^k_\varphi(q) - d^k_\varphi(q') \} \!
            & \!\! \text{if} \, q' \in \Sucq \text{, } q' \! \not \to^* \! q \\
         0 \! & \!\! \text{otherwise }
    \end{cases}
\end{equation}
Then, we define an \emph{adaptive progression reward function} for the $k$-th round of updates as:
\begin{equation} \label{eqn:rap}
    \Rap{k} \left(\langle s, q\rangle, a,\langle s', q'\rangle\right) = 
        \max \{\rho^0_\varphi (q, q'), \rho^k_\varphi (q, q')\} 
\end{equation}
When $k=0$, the adaptive progression reward function $\Rap{0}$ coincides with the progression reward function $\Rp$ defined in \eqnref{eqn:rp}.

\begin{examp} \label{eg:rap}
Using the updated distance-to-acceptance values from \egref{eg:updateD}, we calculate the adaptive progression rewards $\Rap{1}$ for the first round of update. 
For instance, we have 
$\rho^1_\varphi (q_1, q_4) = \max \{0, d^1_\varphi(q_1) - d^1_\varphi(q_4) \} = 101$. 
Recall $\rho^0_\varphi(q_1, q_4)= 1$ from \egref{eg:distance}. 
Thus, 
\[
\Rap{1} \left(\langle g_{6,4}, q_1\rangle, \text{right},\langle g_{6,5}, q_4\rangle\right)
= \max\{1,101\}=101.
\]
The expected returns of policies in \egref{eg:policies} with $\Rap{1}$ are
$\Vap{1}{\pi_1}(s_0^\otimes) \approx 0.39$,
$\Vap{1}{\pi_2}(s_0^\otimes) \approx 13.85$, and 
$\Vap{1}{\pi_3}(s_0^\otimes) = 0$. 
Policy $\pi_2$ yields the highest expected return while completing the task (i.e., $b(\pi_2) = 0$).
\end{examp}

\startpara{Adaptive hybrid reward function}
We define an \emph{adaptive hybrid reward function} for the $k$-th round of updates as:
\begin{align} \label{eqn:rah}
    & \Rah{k} \left(\langle s, q\rangle, a,\langle s', q'\rangle\right) = \nonumber \\
    &\begin{cases}
         \eta_k \cdot - d^k_\varphi(q) & \text { if } q = q' \\
         (1-\eta_k) \cdot \max \{\rho^0_\varphi (q, q'), \rho^k_\varphi (q, q')\} & \text { otherwise }
    \end{cases}
\end{align}
with $\eta_0 \in [0,1]$, and $\eta_k = \frac{\eta_{k-1}}{\theta}$ where $\theta$ is the same hyperparameter used in \eqnref{eqn:updateD}. 
We require $\theta>1$ to ensure that the weight value $\eta_k$ is reduced in each update round, avoiding undesired behavior from increased self-loop penalties.
At $k=0$, the adaptive hybrid reward function $\Rah{0}$ aligns with the hybrid reward function $\Rh$ as defined in \eqnref{eqn:rh}.

\begin{examp} \label{eg:rah}
Let $\eta_0 = 0.1$, and $\theta_1=100$. 
The initial distance-to-acceptance values $d^0_{\varphi}$ are the same as in \egref{eg:updateD}.
Suppose the agent's movement during the episodes follows a policy $\pi$ such that $b(\pi) = 1$.
Following \eqnref{eqn:updateD}, we update the distance-to-acceptance values of states in 
$B_1 \cup B_2 \cup B_3 = \{q_0, q_1, q_2, q_3\}$ to 
$d^1_{\varphi}(q_1) = d^1_{\varphi}(q_2) = 101$,
$d^1_{\varphi}(q_0) = 102$, and $d^1_{\varphi}(q_3) = 115$.
We compute $\Rah{1}$ with $\eta_1 = 0.001$, which yields
$\Vah{1}{\pi_1}(s_0^\otimes) \approx -0.52$,
$\Vah{1}{\pi_2}(s_0^\otimes) \approx 12.97
$,
and $\Vah{1}{\pi_3}(s_0^\otimes) \approx -0.35$.
The optimal policy $\pi_2$ not only yields the highest expected return but also completes the task with $b(\pi_2) = 0$.
\end{examp}


\startpara{Correctness}
The correctness of the proposed adaptive reward shaping approach, as it pertains to the problem formulated in \sectref{sec:problem}, is stated below, with the proof provided in \apref{app:proof}.

\begin{theorem}\label{thm:main}
Given an episodic MDP $\cM$ and a DFA $\cA_\varphi$ corresponding to a co-safe LTL formula $\varphi$, there exists an optimal policy $\pi^*$ of the product MDP $\cM^\otimes = \cM \otimes \cA_\varphi$ that maximizes the expected return based on a reward function $R^{\otimes} \in \{\Rap{k}, \Rah{k}\}$ for some $k \in \Nset$, where the task progression for policy $\pi^*$ matches the best possible task progression $b^*$ across all feasible policies in the product MDP $\cM^\otimes$, that is, $b^* = b(\pi^*)$. 
\end{theorem}

\section{Experiments} \label{sec:exp} 

We evaluate the proposed adaptive reward shaping approach in a variety of benchmark RL domains. 
We describe the experimental setup including environments, RL algorithms, baselines, and evaluation metrics in \sectref{sec:setup}, and analyze the experimental results in \sectref{sec:results}. 

\subsection{Experimental Setup} \label{sec:setup}

\begin{figure*}[t]
    \centering
    \includegraphics[width=0.75\textwidth]{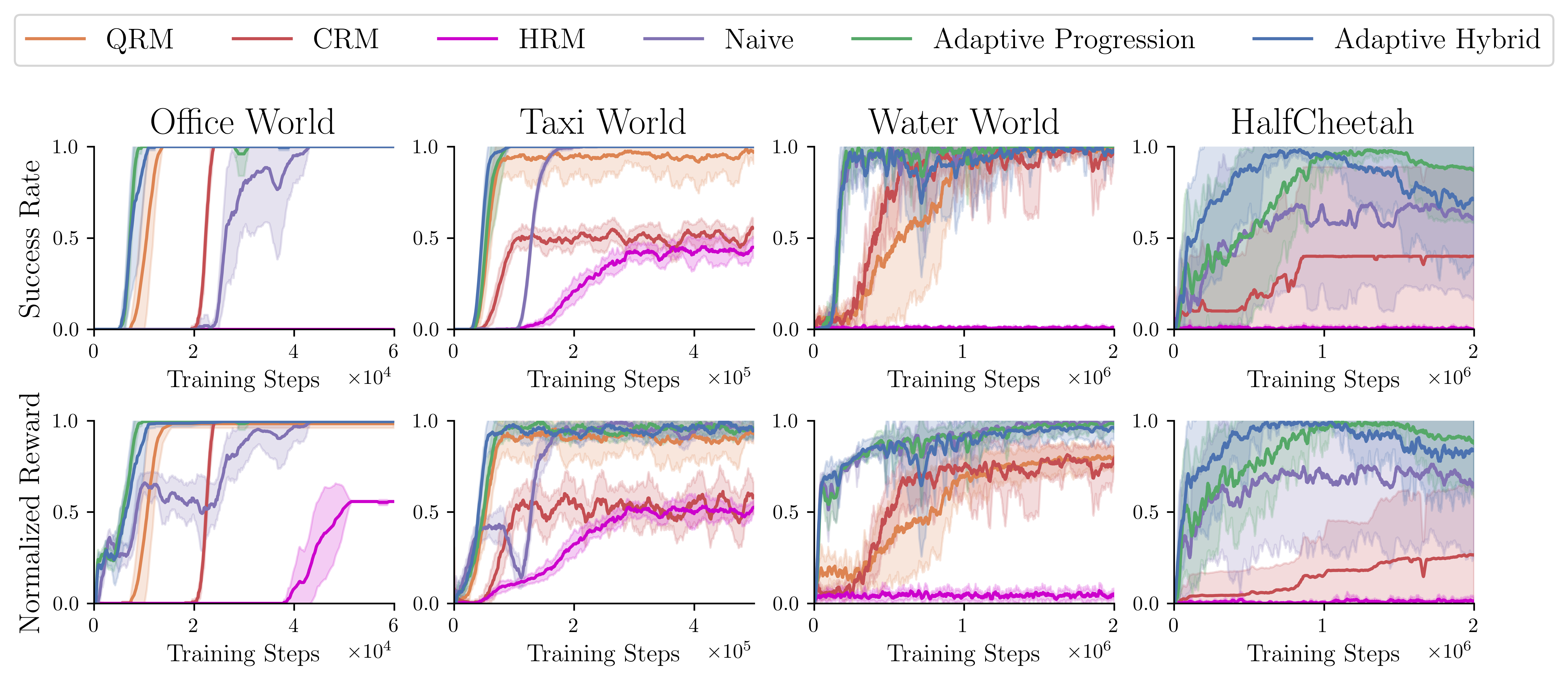}
    \caption{Results for deterministic environments.}
    \label{fig:normal}
\end{figure*}

\begin{figure*}[h]
    \centering
    \includegraphics[width=0.75\textwidth]{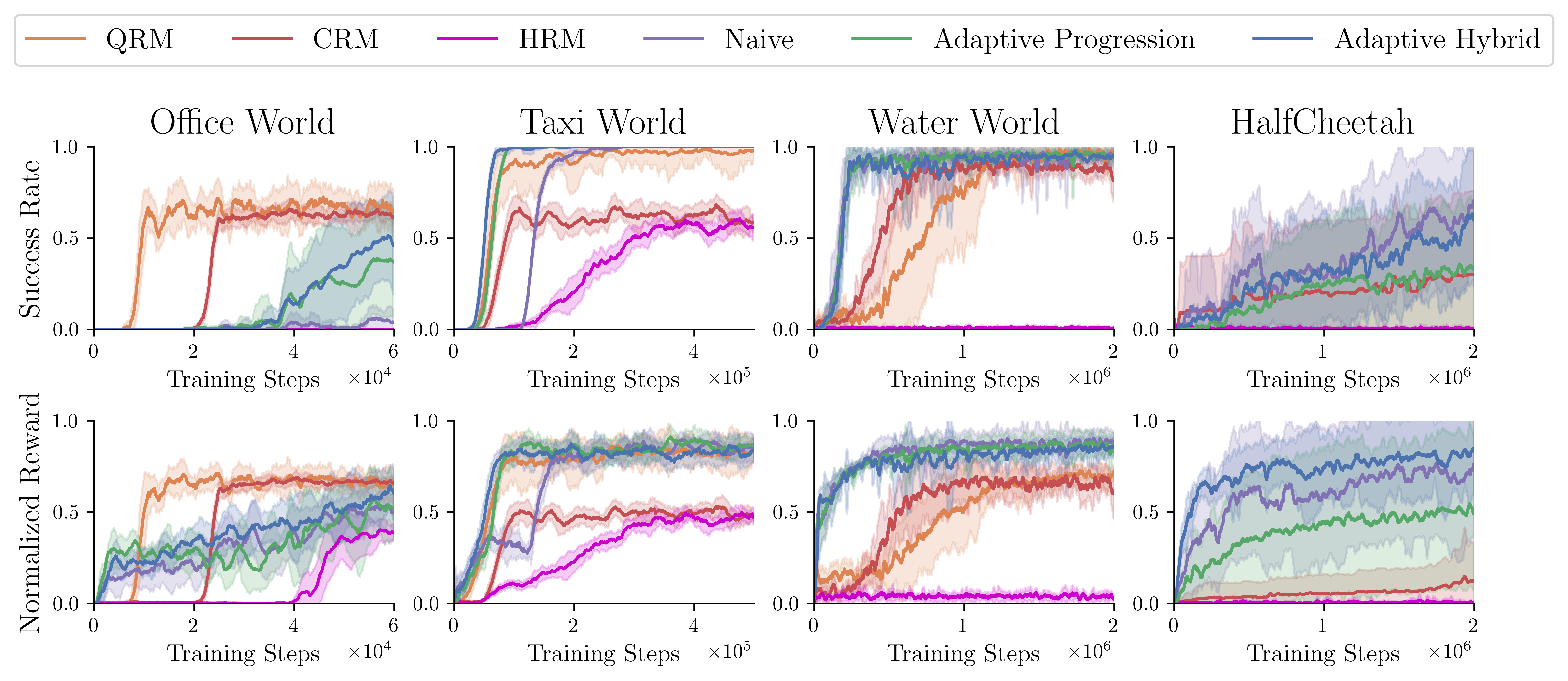}
    \caption{Results for noisy environments.}
    \label{fig:noisy}
\end{figure*}

\startpara{Environments}
The following RL domains are used: the taxi domain from OpenAI Gym~\citep{brockman2016openai}, and three other domains adapted from~\cite{icarte2022reward}.
\begin{itemize}
    \item \emph{Office world}: The agent navigates a 12$\times$9 grid world to: get coffee and mail (in any order), deliver them to the office, and avoid obstacles. The test environment assigns a reward of 1 for each sub-goal: (i) get coffee, (ii) get coffee and mail, and (iii) deliver coffee and mail to the office, all while avoiding obstacles.
    \item \emph{Taxi world}: The agent drives around a 5$\times$5 grid world to pick up and drop off a passenger, starting from a random location. There are five possible pickup locations and four possible destinations. The task is completed when the passenger is dropped off at the target destination. The test environment assigns a reward of 1 for each sub-goal: (i) pick up the passenger, (ii) reach the target destination, and (iii) drop off the passenger. 
    \item \emph{Water world}: The agent moves in a continuous 2D box with six colored floating balls, changing velocity toward one of the four cardinal directions each step. The task is to touch red and green balls in strict order without touching other colors, then touch blue balls. The test environment assigns a reward of 1 for touching each target ball.
    \item \emph{HalfCheetah}: The agent is a cheetah-like robot with a continuous action space, controlling six joints to move. The task is completed by reaching the farthest location. The test environment assigns a reward of 1 for reaching each of the five locations along the way.
\end{itemize}
For each domain, we consider three types of environments: 
(1) \emph{deterministic} environments, where each state-action pair leads to a single success state only; 
(2) \emph{noisy} environments, where each action has a certain control noise; 
and (3) \emph{infeasible} environments, where some sub-goals are impossible to complete (e.g., a blocked office that the agent cannot access, or missing blue balls in the water world).

\startpara{Baselines}
We compare the proposed approach with the following methods as baselines: 
\emph{Q-learning for reward machines} (QRM) with reward shaping~\citep{camacho2019ltl},  \emph{counterfactual experiences for reward machines} (CRM) with reward shaping and \emph{hierarchical RL for reward machines} (HRM) with reward shaping~\citep{icarte2022reward}. We also evaluate RM-based algorithms incorporating partial rewards, which are detailed in Appendix~\ref{app:pr rm}. We use the code accompanying publications. 

Moreover, we consider a naive baseline that rewards transitions that decrease the distance to acceptance. For each transition $\left(\langle s, q\rangle, a, \langle s', q'\rangle\right)$ in the product MDP, assign a reward of 1 if $d_\varphi(q) > d_\varphi(q')$ and there is a path from $q$ to accepting states $Q_F$, otherwise assign a reward of 0.


\startpara{RL Algorithms} 
We use DQN~\cite{dqn2015} for learning in discrete domains (office world and taxi world), DDQN~\citep{ddqn2016} for water world with continuous state space, and DDPG~\citep{ddpg2016} for HalfCheetah with continuous action space. 
Note that QRM implementation does not work with DDPG, so we only use HRM and CRM as the baselines for HalfCheetah. 
We also apply PPO~\citep{ppo2017} and A2C~\citep{a2c2016} to HalfCheetah (QRM, CRM and HRM baselines are not compatible with these RL algorithms) and report results in \apref{app:cheetah} due to the page limit. 
Our implementation was built upon OpenAI Stable-Baselines3~\citep{stable-baselines3}.

\startpara{Metrics}
We pause the learning process every 100 training steps in the office world and every 1,000 training steps in other domains, then evaluate the current policy in the test environment over 5 episodes.
We evaluate the performance using two metrics: \emph{success rate of task completion}, calculated by counting the frequency of successful episodes where the task is completed, and \emph{normalized expected return}, which is normalized using the maximum possible return for that task.
The only exception is taxi world, where the maximum return varies for different initial states and we normalize by averaging the maximum return of all initial states.

\subsection{Results Analysis} \label{sec:results}

\begin{figure*}[t]
    \centering
    \includegraphics[width=0.75\textwidth]{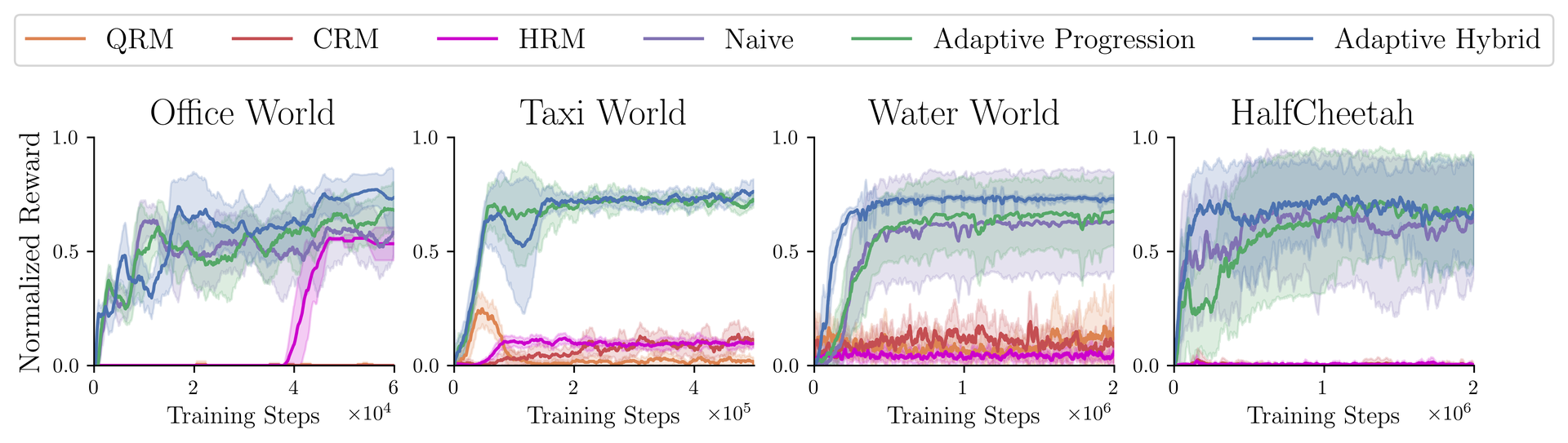}
    \caption{Results for infeasible environments.}
    \label{fig:missing}
\end{figure*}

We ran 10 independent trials for each method. 
\figfigfigref{fig:normal}{fig:noisy}{fig:missing} plot the mean performance with a $95\%$ confidence interval (the shaded area) in deterministic, noisy, and infeasible environments, respectively. 
The success rate of task completion is omitted in \figref{fig:missing} because it is zero for all trials (i.e., the task is infeasible to complete).

\startpara{Performance comparison}
These results show that the proposed approach using adaptive progression or adaptive hybrid reward functions generally outperforms baselines, achieving earlier convergence to policies with a higher success rate of task completion and a higher normalized expected return. 

The significant advantage of our approach is best illustrated in \figref{fig:missing}, where baselines fail to learn effectively in environments with infeasible tasks.
Although baselines apply potential-based reward shaping~\citep{ng1999policy} to assign intermediate rewards, they cannot distinguish between good and bad terminal states (e.g., completing a sub-goal and colliding with an obstacle have the same potential value).
In contrast, our approach provides more effective intermediate rewards, encouraging the agent to learn and maximize task progression.

The only outlier is the noisy office world where QRM and CRM outperform the proposed approach. 
One possible reason is that our approach gets stuck with a sub-optimal policy in this environment, which opts for fetching coffee at a closer location but results in a longer route to complete other sub-goals (i.e., getting mail and delivering to office).

Comparing the proposed reward functions, we observe that the adaptive hybrid reward function has the best overall performance. 
Comparing different RL environments, the proposed approach can achieve a success rate of 1 and the maximum possible return in most deterministic environments, but its performance is degraded in noisy environments due to control noise and in infeasible environments due to environmental constraints.

\begin{figure}[h]
    \centering
    \includegraphics[width=\columnwidth]{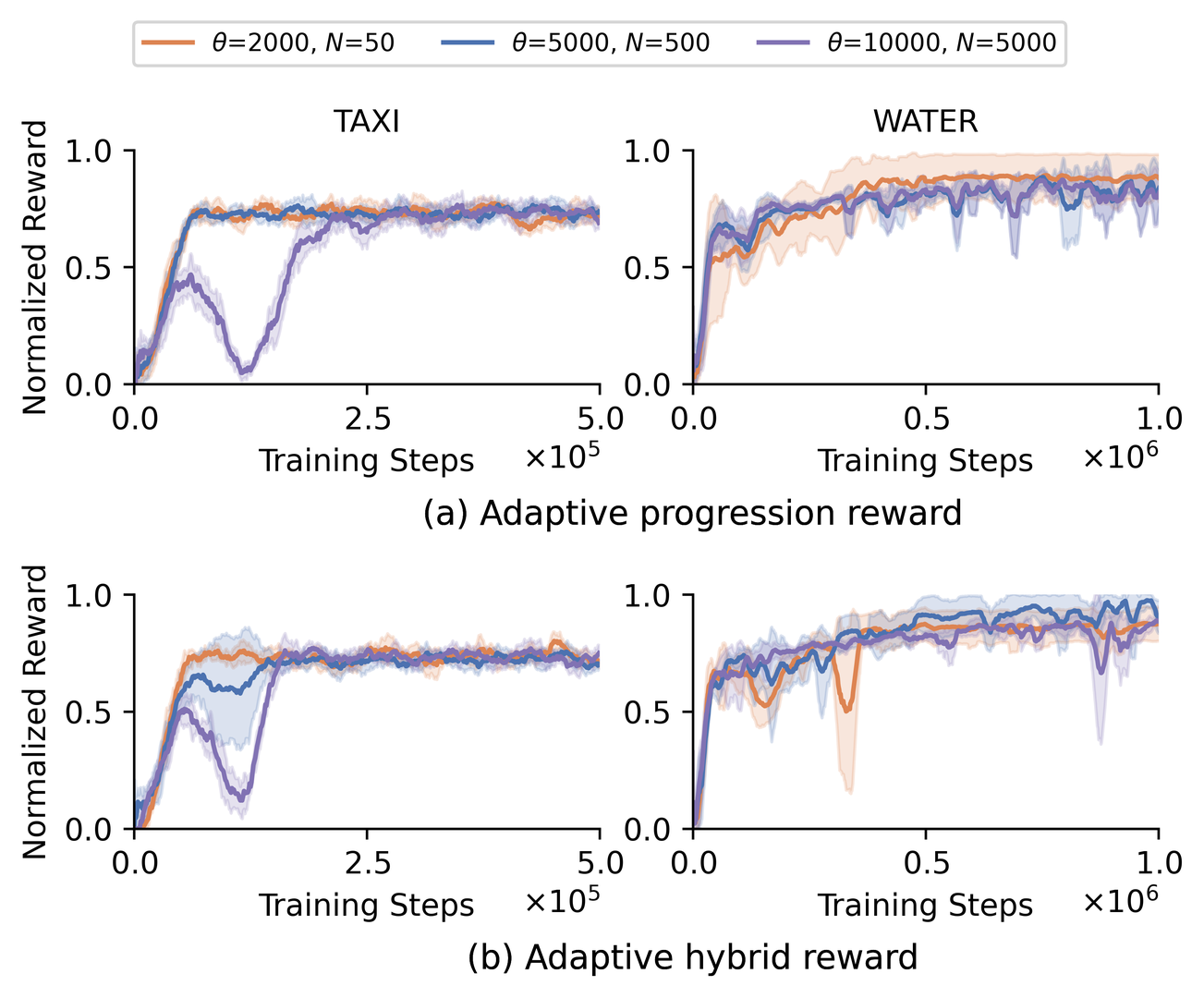}
    \caption{Results of the ablation study on the sensitivity of hyperparameters $\theta$ and $N$ for updating distance-to-acceptance values in infeasible environments.} 
    \label{fig:ablation1}
\end{figure}

\begin{figure}[h]
    \centering
    \includegraphics[width=\columnwidth]{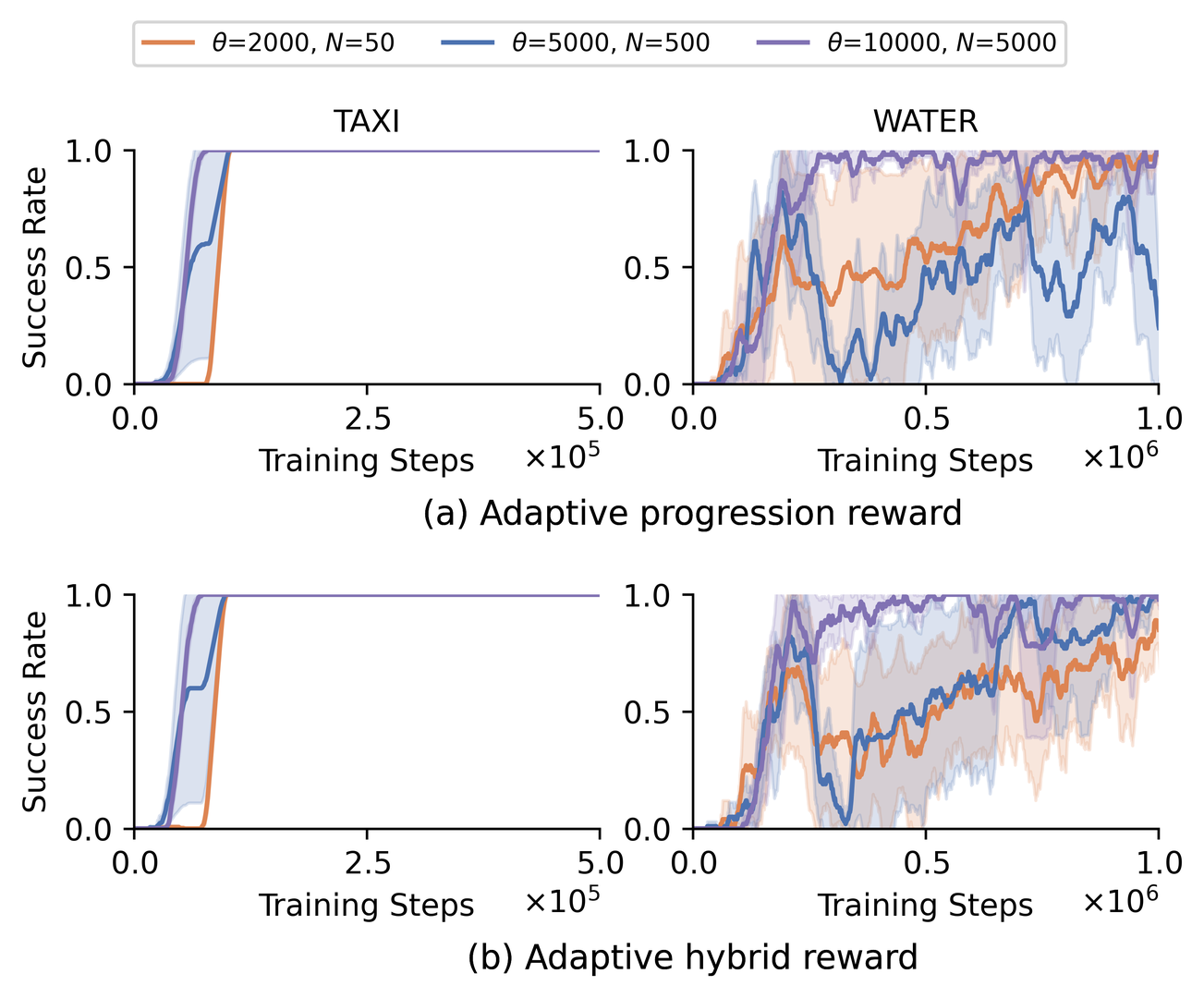}
    \caption{Results of the ablation study on the sensitivity of hyperparameters $\theta$ and $N$ for updating distance-to-acceptance values in feasible environments.} 
    \label{fig:ablation2}
\end{figure}

\startpara{Ablation study}
Additionally, we conduct an ablation study to investigate the sensitivity of the hyperparameters $\theta$ and $N$ used for updating distance-to-acceptance values (cf. \sectref{sec:adapt}). 
\figref{fig:ablation1} shows the normalized reward for two infeasible environments: Taxi World and Water World. 
The results demonstrate that the proposed approach converges with a sufficiently large value of $\theta \in \{2,000, 5,000, 10,000\}$. Moreover, it takes more training steps to achieve policy convergence with larger values of $N$, indicating longer intervals between consecutive updates of reward functions.
\figref{fig:ablation2} shows the success rates for the feasible version of Taxi World and Water World. These results suggest that feasible environments benefit from longer update intervals $N$, as the agent has more time to explore and gather experience before the reward function is modified.

\startpara{Hyperparameter Selection: Practical Guidance}
We offer the following heuristics for selecting key hyperparameters in our framework. For the reward update interval $N$, a useful starting point is the total training budget divided by the number of distinct task stages (e.g., states in a task-governing DFA), as this aims to provide the agent with sufficient interaction episodes within each task stage before potential reward adjustments. The reward scaling factor $\theta$ can be initially set to the sum of progression rewards, $\sum_{q, q^{\prime}} p_\phi\left(q, q^{\prime}\right)$, which approximates the cumulative effort. Insights from the ablation study further suggest that task feasibility can guide these choices: potentially infeasible tasks may benefit from smaller $\theta$ values and more frequent updates (smaller $N$) to enable regular progress assessment and dynamic reward adjustment in challenging settings. In contrast, feasible tasks often accommodate larger $\theta$ and $N$ to allow for the collection of more meaningful data before reward function modification. When employing hybrid reward functions, we recommend small magnitudes for step-wise penalties (e.g., $10^{-3}$ or $10^{-4}$) to avoid overwhelming the positive shaping signals. These guidelines serve as practical starting points, though optimal settings can be task-dependent. 

\section{Conclusion} \label{sec:conclu} 
We have developed a logic-based adaptive reward shaping approach for RL. Our approach uses reward functions designed to incentivize an agent to complete a task specified by a co-safe LTL formula as much as possible, and dynamically updates these reward functions during the learning process. This dynamic reward shaping is beneficial for scenarios where environmental uncertainties can lead to task failure despite successful subtask progress.

Computational experiments demonstrate that our approach is applicable to various discrete and continuous RL domains and is compatible with a wide range of RL algorithms such as DQN, DDQN, DDPG, PPO, and A2C. Experimental results also show that the proposed approach generally outperforms state-of-the-art baselines, achieving faster convergence to a better policy with higher expected return and task completion rate.

There are several directions for future work. First, we will evaluate the proposed approach on a broader range of RL domains beyond the benchmarks used in our experiments. Second, we will explore extending the approach to multi-agent RL. Finally, we aim to apply the proposed approach to real-world RL tasks, such as autonomous driving.

\section*{Acknowledgments}

This work was supported in part by the U.S. National Science Foundation under Grants CCF-1942836 and CCF-2131511. The opinions, findings, conclusions, or recommendations expressed in this material are those of the author(s) and do not necessarily reflect the views of the sponsoring agencies.


\bibliography{references}

\newboolean{arxiv}
\setboolean{arxiv}{true} 

\newpage

\onecolumn

\title{Supplementary Material}
\maketitle

\appendix
\setcounter{section}{0}
\renewcommand{\thesection}{\Alph{section}}
\section{Correctness} \label{app:proof}
Here, we prove the correctness of our approach, as stated in \thmref{thm:main}. 
We start by proving the following auxiliary lemmas.

\begin{lemma}\label{lem:h2p}
Adaptive hybrid reward function $\Rah{k}$ tends to adaptive progression reward function $\Rap{k}$ with an increasing number of updates $k$, that is, $\lim_{k \to \infty} \Rah{k} = \Rap{k}$. 
\end{lemma}

\begin{proof}
By the definition of adaptive hybrid reward function $\Rah{k}$ (cf. \eqnref{eqn:rah}), $\eta_0 \in [0,1]$ and $\eta_k = \frac{\eta_{k-1}}{\theta}$ with $\theta > 1$.
We have $\lim_{k \to \infty} \eta_k = 0$. 
The first case of \eqnref{eqn:rah}, $\eta_k \cdot - d^k_\varphi(q)$, tends to $0$; and the second case tends to $\max \{\rho^0_\varphi (q, q'), \rho^k_\varphi (q, q')\}$.
Thus, it holds that $\lim_{k \to \infty} \Rah{k} = \Rap{k}$.
\end{proof}

\begin{lemma}\label{lem:progress}
Given an episodic MDP $\cM$ and a DFA $\cA_\varphi$ for a co-safe LTL formula $\varphi$, let $\pi^{*}_k$ and $\pi^{*}_{k+1}$ denote the optimal policies of the product MDP $\cM^\otimes = \cM \otimes \cA_\varphi$, maximizing the expected return based on adaptive progression reward functions $\Rap{k}$ and $\Rap{k+1}$, respectively. If a policy exists that achieves a higher expected return than $\pi^{*}_k$ based on $\Rap{k+1}$, then $\pi^{*}_{k+1}$ achieves better task progression than $\pi^{*}_k$, that is, $b(\pi^{*}_{k+1}) < b(\pi^{*}_k)$. 
\end{lemma}

\begin{proof}
For the sake of contradiction, suppose that $b(\pi^{*}_{k+1}) \ge b(\pi^{*}_k)$. 
Let $\tau$ be a path through the product MDP $\cM^\otimes$ under policy $\pi^{*}_{k+1}$. 
For any state $\langle s, q\rangle$ in the path $\tau$, we have $q \in B_i$ where $i \ge b(\pi^{*}_{k+1}) \ge b(\pi^{*}_k) = b_k$.
For every transition $(\langle s, q\rangle, a,\langle s', q'\rangle) \in \tau$, it holds that:
\begin{align*} 
& \Rap{k+1} \left(\langle s, q\rangle, a,\langle s', q'\rangle\right) \\
    =& \max \{\rho^0_\varphi (q, q'), \rho^{k+1}_\varphi (q, q')\} \\
    =& \max \{\rho^0_\varphi (q, q'), \max \{0, d^{k+1}_\varphi(q) - d^{k+1}_\varphi(q') \} \} \\
    =& \max \{\rho^0_\varphi (q, q'), \max \{0, d_\varphi^{k}(q) + \theta - d_\varphi^{k}(q') - \theta \} \} \\
    =& \max \{\rho^0_\varphi (q, q'), \max \{0, d^k_\varphi(q) - d^k_\varphi(q') \} \} \\   
    =& \max \{\rho^0_\varphi (q, q'), \rho^k_\varphi (q, q') \} \\
    =& \Rap{k} \left(\langle s, q\rangle, a,\langle s', q'\rangle\right)   
\end{align*}
Thus, we have $\Vap{k+1}{\pi^{*}_{k+1}}(s_0^{\otimes}) = \Vap{k}{\pi^{*}_{k+1}}(s_0^{\otimes})$, meaning that the expected return stays the same.
Similarly, we can show that $\Vap{k+1}{\pi^{*}_{k}}(s_0^{\otimes}) = \Vap{k}{\pi^{*}_{k}}(s_0^{\otimes})$.

Since $\pi^{*}_{k}$ is the optimal policy maximizing the expected return based on $\Rap{k}$, we have 
\begin{equation} \label{eqn:c1}
    \Vap{k}{\pi^{*}_{k}}(s_0^{\otimes}) \ge \Vap{k}{\pi^{*}_{k+1}}(s_0^{\otimes}) = 
    \Vap{k+1}{\pi^{*}_{k+1}}(s_0^{\otimes}).
\end{equation}

Given that there exists a policy that achieves a higher expected return than $\pi^{*}_k$ based on $\Rap{k+1}$, 
it holds that
\begin{equation} \label{eqn:c2}
    \Vap{k}{\pi^{*}_{k}}(s_0^{\otimes}) = \Vap{k+1}{\pi^{*}_k}(s_0^{\otimes}) < 
    \Vap{k+1}{\pi^{*}_{k+1}}(s_0^{\otimes}).
\end{equation}

\eqnref{eqn:c1} contradicts with \eqnref{eqn:c2}. Thus, we have $b(\pi^{*}_{k+1}) < b(\pi^{*}_k)$. 
\end{proof}

Now we are ready to prove \thmref{thm:main} as stated in \sectref{sec:approach} and repeated here. 
\setcounter{theorem}{0}
\begin{theorem}
Given an episodic MDP $\cM$ and a DFA $\cA_\varphi$ corresponding to a co-safe LTL formula $\varphi$, there exists an optimal policy $\pi^*$ of the product MDP $\cM^\otimes = \cM \otimes \cA_\varphi$ that maximizes the expected return based on a reward function $R^{\otimes} \in \{\Rap{k}, \Rah{k}\}$ for some $k \in \Nset$, where the task progression for policy $\pi^*$ matches the best possible task progression $b^*$ across all feasible policies in the product MDP $\cM^\otimes$, that is, $b^* = b(\pi^*)$. 
\end{theorem}

\begin{proof}
Without loss of generality, we focus on the adaptive progression reward function $\Rap{k}$, as \lemref{lem:h2p} shows that $\lim_{k \to \infty} \Rah{k} = \Rap{k}$. 

Let $\pi^{*}_k$ denote an optimal policy of the product MDP $\cM^\otimes$ that maximizes the expected return based on $\Rap{k}$.
Suppose that $b(\pi^{*}_k) > b^*$. 
There exists a policy $\pi$ in the product MDP that achieves the best possible task progression $b^*$, where 
$\Vap{k}{\pi}(s_0^{\otimes}) \le \Vap{k}{\pi^{*}_{k}}(s_0^{\otimes})$.
If $\Vap{k}{\pi}(s_0^{\otimes}) = \Vap{k}{\pi^{*}_{k}}(s_0^{\otimes})$, then $\pi$ is the desired optimal policy $\pi^*$ that maximizes the expected return based on $\Rap{k} $while achieving the best possible task progression $b^*$. This theorem is thus proved. 

Otherwise, when $\Vap{k}{\pi}(s_0^{\otimes}) < \Vap{k}{\pi^{*}_{k}}(s_0^{\otimes})$, we proceed to prove the theorem as follows. 
Let the difference in expected returns be denoted by 
$\sigma = \Vap{k}{\pi^{*}_{k}}(s_0^{\otimes}) - \Vap{k}{\pi}(s_0^{\otimes}) > 0$.
Consider a worst-case scenario where policy $\pi$ reaches a state with the best possible task progression only at the end of an episode. 
Formally, there is only one path $\tau$ of length $|\tau|=H$ through the product MDP $\cM^\otimes$ under policy $\pi$ that ends with a transition $(\langle s, q\rangle, a, \langle s', q'\rangle)$ where $q \in B_i$, $q' \in B_j$, and $i > j = b^*$.
Based on the definition of adaptive progression reward function, we have 
$\Rap{k+1} \left(\langle s, q\rangle, a,\langle s', q'\rangle\right) = \Rap{k} \left(\langle s, q\rangle, a,\langle s', q'\rangle\right) + \theta$. 
Thus, $\Vap{k+1}{\pi}(s_0^{\otimes}) = \Vap{k}{\pi}(s_0^{\otimes}) + p \cdot \gamma^{H-1} \cdot \theta$, 
where $p$ is the probability of path $\tau$ and $\gamma$ is the MDP's discount factor. 
Following the argument in \lemref{lem:progress}, it holds that 
$\Vap{k+1}{\pi^{*}_{k}}(s_0^{\otimes}) = \Vap{k}{\pi^{*}_{k}}(s_0^{\otimes})$. 
When the hyperparameter value $\theta$ is sufficiently large, more precisely,
$\theta > \frac{\sigma}{p \cdot \gamma^{H-1}}$,
we have $\Vap{k+1}{\pi}(s_0^{\otimes}) > \Vap{k+1}{\pi^{*}_{k}}(s_0^{\otimes})$. 
Let $\pi^{*}_{k+1}$ denote an optimal policy of the product MDP $\cM^\otimes$ that maximizes the expected return based on $\Rap{k+1}$.
If $\Vap{k+1}{\pi}(s_0^{\otimes}) = \Vap{k+1}{\pi^{*}_{k+1}}(s_0^{\otimes})$, then $\pi$ is the desired optimal policy $\pi^*$ and the theorem is thus proved. 
Otherwise, following \lemref{lem:progress}, it holds that $b(\pi^{*}_{k+1}) < b(\pi^{*}_k)$, meaning that the task progression for $\pi^{*}_{k+1}$ has improved compared to that of policy $\pi^{*}_k$. 
Since a task progression value is bounded by the state partition size of DFA $\cA_\varphi$, it takes only a finite number of updates before an optimal policy yielding $b^*$ is learned.

In conclusion, there exists an optimal policy $\pi^*$ for the product MDP $\cM^\otimes$ that achieves the best possible task progression $b^*$ while maximizing the expected return based on $\Rap{k}$ for some $k \in \Nset$, which is an adaptive progression reward function updated in a finite number of rounds with a sufficiently large hyperparameter value $\theta$.
\end{proof}
\newpage
\section{Compatibility with On-Policy Learning}\label{app:cheetah}
\begin{figure}[h]
    \centering
    \includegraphics[width=\columnwidth]{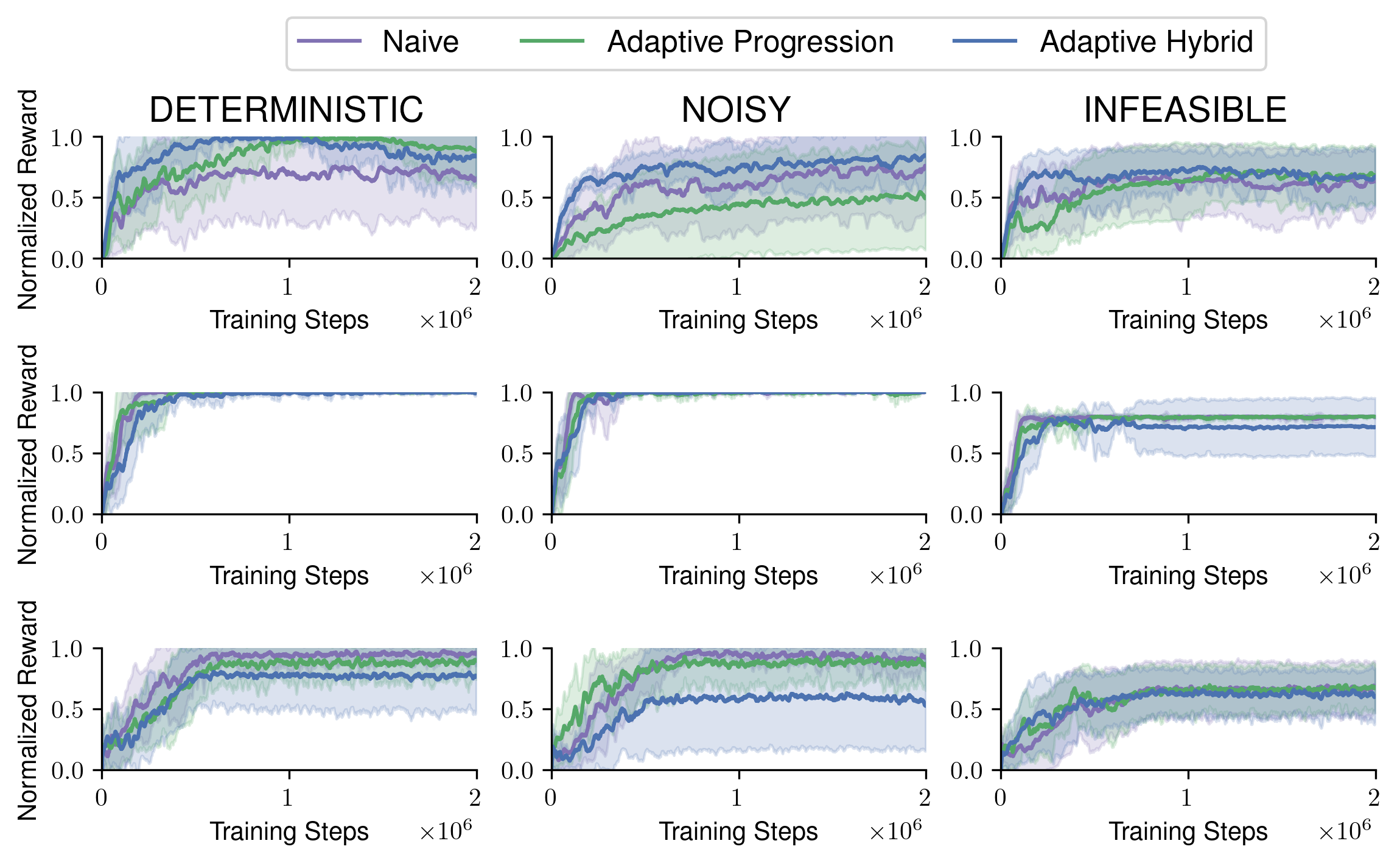}
    \caption{Results of applying various RL algorithms to HalfCheetah environments.}   
    \label{fig:cheetah}
\end{figure}

\startpara{Results for HalfCheetah} 
\figref{fig:cheetah} shows the results of applying three different RL algorithms, DDPG~\citep{ddpg2016}, PPO~\citep{ppo2017}, and A2C~\citep{a2c2016}, to HalfCheetah environments.
The comparison between the proposed approach and all baselines using DDPG has already been discussed in \sectref{sec:exp}.
Since the QRM, CRM, and HRM baselines are not compatible with PPO and A2C, we only compare with the naive baseline here.

Comparing the results of the three RL algorithms, we observe that DDPG exhibits relatively higher variance than the others. This is likely due to its off-policy nature, relying heavily on a replay buffer and exploration driven by control noise. In our experiments, we used a replay buffer with a capacity of $10^6$ while sampling only $100$ experiences for each update, introducing significant randomness as most samples in the large replay buffer do not yield positive rewards. Exploration also adds to the randomness.
In contrast, PPO and A2C are on-policy algorithms, where updates depend solely on the current policy. These algorithms tend to maintain their behavior once the current policy achieves partial task completion. Additionally, PPO incorporates a stabilizing technique that helps reduce variance.

Comparing different reward functions, we find that the Naive baseline achieves comparable performance with the proposed reward functions in all HalfCheetah environments. However, as noted in \sectref{sec:exp}, it usually performs the worst in other domains. One possible explanation is that the HalfCheetah task has a unique structure, where each sub-goal requires moving forward by the same distance. The Naive reward function assigns a reward of 1 for each sub-goal, maintaining consistency in the learning process.

\newpage
\section{Partial Rewards in Reward Machines}\label{app:pr rm}
This ablation study investigates the effect of incorporating partial rewards into Reward Machine (RM) structures along with the use of potential-based reward shaping. While RMs are theoretically capable of representing and utilizing partial rewards (e.g., in the Office World environment, the RM transitions through states $u_0$ [initial], $u_1, u_2$ [intermediate], and $u_3$ [goal], as depicted in Figure 2 (b)~\cite{icarte2022reward}, our empirical evaluation reveals that their inclusion does not consistently enhance performance and can lead to performance degradation.

To evaluate the impact of partial rewards on RM-based algorithms, we conducted experiments in deterministic environments: Office World and Taxi World. We define "Partial Reward Q-learning Reward Machine" (PR QRM) as the QRM algorithm variant that incorporates partial rewards. For consistency across algorithms, we similarly introduce PR CRM and PR HRM, denoting CRM and HRM variants also utilizing partial rewards. All baselines in this study are equipped with potential-based reward shaping. Across both Office World and Taxi World environments, all RM-based algorithms suffered a performance degradation when supplemented with partial rewards of $1$ for each intermediate step.

These findings suggest that the algorithms, particularly in their current configurations, may not be inherently designed to effectively leverage partial rewards in conjunction with potential reward shaping. One plausible explanation for the observed performance degradation is that, as discussed in~\cite{icarte2022reward}, potential reward shaping can assign positive rewards to actions that lead to undesirable "violation" states within the RM, potentially exacerbating the negative effects of partial rewards. Therefore, careful consideration and potentially algorithm modifications are necessary to effectively harness the benefits of partial rewards within Reward Machine frameworks, especially when integrated with reward shaping techniques. In contrast to these observations, our proposed algorithms are designed to effectively incorporate partial rewards across diverse environments without performance degradation, while ensuring both task completion and reward maximization.

\begin{figure}[h]
    \centering
    \includegraphics[width=\columnwidth]{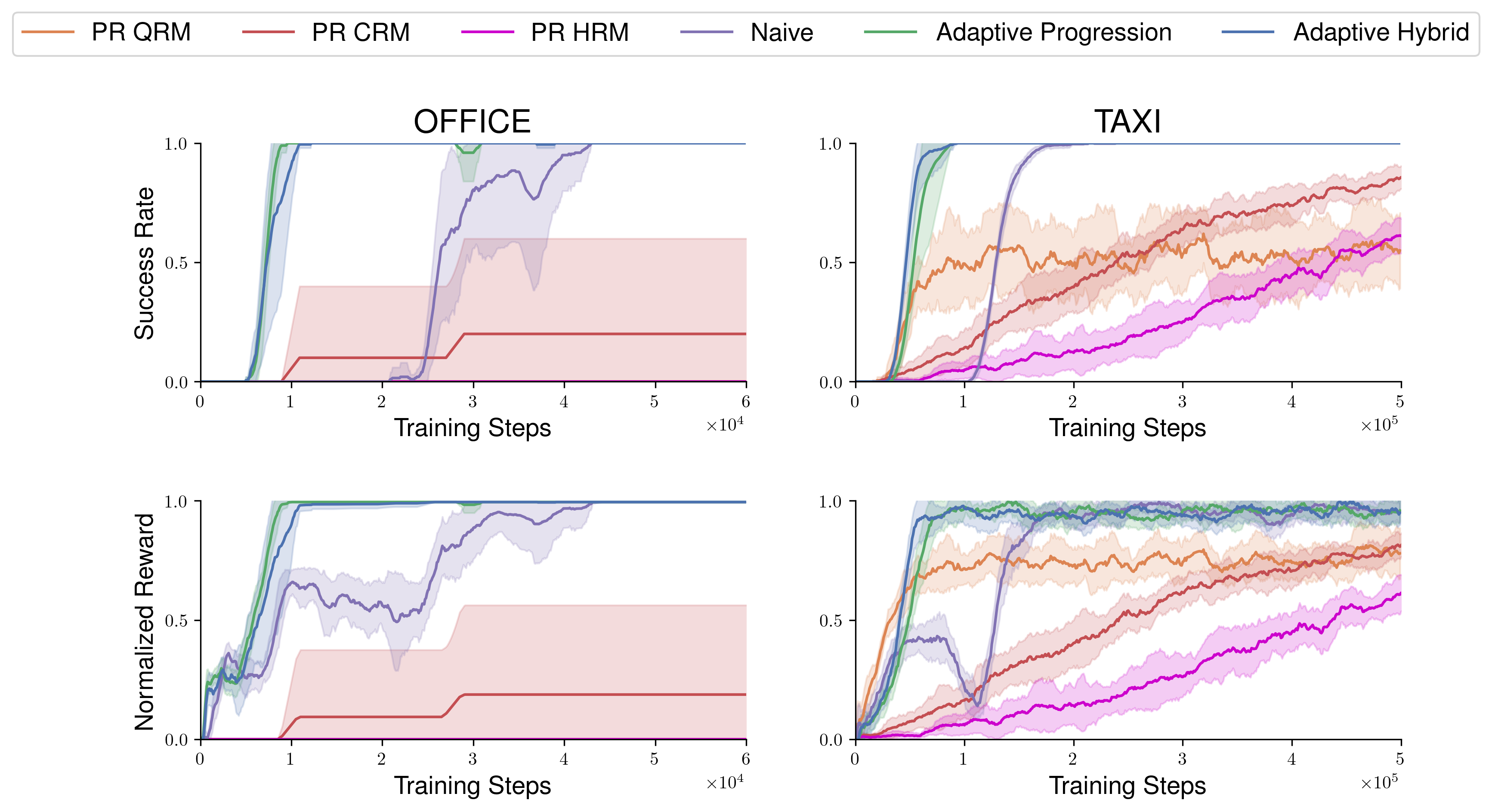}
    \caption{Results in Office World and Taxi World for RM algorithms using partial rewards.}   
    \label{fig:PR RM}
\end{figure}

\end{document}